\documentclass[10pt,twocolumn,letterpaper]{article}
\usepackage{iccv}
\usepackage{times}
\usepackage{epsfig}
\usepackage{graphicx}
\usepackage{amsmath}
\usepackage{amssymb}
\usepackage{float}
\usepackage{enumitem}
\usepackage{subcaption}
\usepackage{xcolor}
\usepackage{authblk}

\usepackage[pagebackref=true,breaklinks=true,letterpaper=true,colorlinks,bookmarks=false]{hyperref}

\usepackage{umoline}

\newcommand{\myparagraph}[1]{\vspace{3pt}\noindent{\bf #1}}

\iccvfinalcopy 

\begin{document}
\title{Interpreting Adversarial Examples with Attributes \vspace{-3mm}}

\author[1]{Sadaf Gulshad}
\author[2]{Jan Hendrik Metzen}
\author[1]{Arnold Smeulders}
\author[1]{Zeynep Akata}
\affil[1]{UvA-Bosch Delta Lab\\
University of Amsterdam, The Netherlands\\}
\affil[2]{Bosch Center for AI (BCAI), Renningen, Germany}
\maketitle

\begin{abstract}
   Deep computer vision systems being vulnerable to imperceptible and carefully crafted noise have raised questions regarding the robustness of their decisions. We take a step back and approach this problem from an orthogonal direction. We propose to enable black-box neural networks to justify their reasoning both for clean and for adversarial examples by leveraging attributes, i.e. visually discriminative properties of objects. We rank attributes based on their class relevance, i.e. how the classification decision changes when the input is visually slightly perturbed, as well as image relevance, i.e. how well the attributes can be localized on both clean and perturbed images. We present comprehensive experiments for attribute prediction, adversarial example generation, adversarially robust learning, and their qualitative and quantitative analysis using predicted attributes on three benchmark datasets.
\end{abstract}

\section{Introduction}
Deep neural networks, despite their good performance in classification~\cite{krizhevsky2012imagenet, ren2015faster, he2017mask, silver2017mastering}, can be easily fooled by adversarial examples, i.e. added imperceptible noise not visible to humans~\cite{szegedy2013intriguing, dong2017towards, carlini2019evaluating, su2018robustness}. Understanding why this happens is of major curiosity. Previous research has provided insights about deep learning frameworks \cite{su2018robustness}, the geometry of their class boundaries ~\cite{szegedy2013intriguing, moosavi2018robustness, 43405} and the geometry of data manifold \cite{gilmer2018adversarial} which have led to a number of detection and defense methods ~\cite{meng2017magnet,metzen2017detecting,tramer2017ensemble}. However, none of them have yet succeeded completely to rectify, detect, or create a defense against adversarial examples. In this work, we take a step back and propose to understand why a deep network is fooled by adversarial examples.

\begin{figure}[t]
        \includegraphics[width=\linewidth, trim=0 0 0 0, clip]{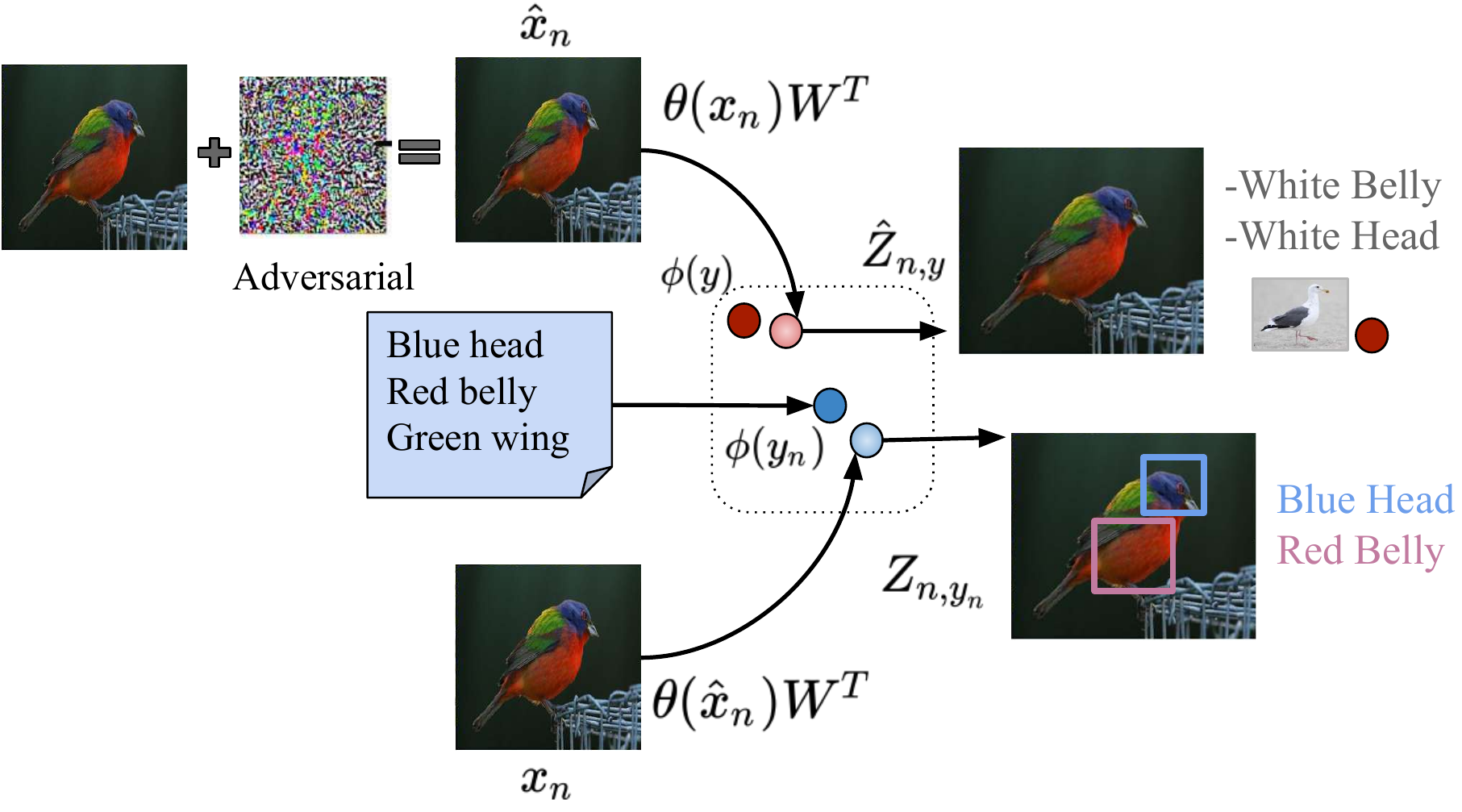}
    \caption{Our interpretable attribute prediction-grounding framework provides visual evidence for a clean image that gets embedded close to the correct blue class (painted bunting) because of ``red belly'' and ``blue head'' attributes, and for an adversarial image that gets embedded close to the incorrect red class (herring gull) as the network thought there were ``white belly'' and ``white head'' attributes.}
    \label{fig:Motivation}
    \vspace{-3mm}
\end{figure}
        
Interpreting deep neural network decisions helps in understanding their internal functioning and could be used for detecting and creating defenses against adversarial attacks \cite{tao2018attacks,du2018towards}. This indirectly provides a way to revisit the decision maker in its failure mode~\cite{tsipras2018robustness}. Previously instance level visual interpretations, e.g. either adding perturbations to the input or by taking the gradient of output with respect to its input \cite{selvaraju2017grad,fong2017interpretable}, have been used to introspect deep neural networks. However, \cite{zhang2018interpretable,adebayo2018sanity} showed that they do not accurately capture the attacks to input generation process and models (see a visual example in Figure~\ref{fig:saliency}).

In this paper, we propose an alternative visual interpretation technique using visually discriminative properties of the objects, i.e. attributes, that are predicted and grounded on clean and adversarial examples. To predict attributes, we learn a mapping from image feature space into class attribute space. Thanks to the ranking based learning, we observe that clean images get mapped close to the correct class while adversarial images get mapped closer to a wrong class embedding. For instance, as shown in Figure~\ref{fig:Motivation}, ``blue head'' and ''red belly'' associated with the class ``painted bunting'' are predicted correctly for the clean image. On the other hand, due to predicting attributes incorrectly as ``white belly'' and ``white head'', the adversarial image gets classified into ``herring gull'' incorrectly. Note that, we consider adversarial examples that are generated to fool only the classifier and not the interpretation mechanism. To ground attributes, we adapt state of the art deep object/object part detector, i.e. Faster-RCNN, to detect bounding boxes around the visual evidence of our predicted attributes. Finally, our analysis involves studying adversarially robust models, i.e. using adversarial training as a defense technique against adversarial attacks.  

Our main contributions are as follows: (1) We propose to understand the neural network decisions for adversarial examples by learning to predict visually discriminative class-specific attributes. (2) We visualize the predicted attributes by grounding them on their respective images, i.e. drawing bounding boxes around their visual evidence on the image. (3) We interpret adversarial examples of standard and adversarially robust framework in three benchmark attribute datasets with varying size and granularity.

\section{Related Work}
In this section, we discuss related works on adversarial examples and interpretability research prior to ours.

\myparagraph{Adversarial Examples.} Small carefully crafted perturbations, i.e. \textit{adversarial perturbations}, added to the inputs of deep neural networks, i.e. \textit{adversarial examples}, can easily fool the classifiers trained using deep learning~\cite{szegedy2013intriguing}. Such attacks involve iterative fast gradient sign method \cite{kurakin2016adversarial}, Jacobian-based saliency map attacks \cite{papernot2016limitations}, one pixel attacks \cite{su2019one}, Carlini and Wagner attacks \cite{carlini2017towards} and universal attacks \cite{moosavi2016deepfool} designed not only for classificaton but for object detection \cite{xie2017adversarial}, segmentation \cite{fischer2017adversarial}, auto encoders \cite{tabacof2016adversarial}, generative models \cite{kos2018adversarial}, and reinforcement learning \cite{lin2017tactics}.  Most of these perturbations are transferable between different networks and do not require access to the network's architecture or parameters, i.e. \textit{black box attacks}.

Concurrently many attempts have been made for understanding, detecting and defense against these attacks. The reason behind adversarial examples may be the linearity in neural networks~\cite{szegedy2013intriguing} or low probability adversarial pockets in image space~\cite{43405}. Neural networks respond to recurrent discriminative patches~\cite{dong2017towards} whereas adversarial examples lie in a different region on the data manifold \cite{gilmer2018adversarial}. Hence, several methods have been proposed to detect adversarial examples~\cite{meng2017magnet,metzen2017detecting}. On the other hand, ensemble adversarial training \cite{tramer2017ensemble},deep contractive networks \cite{gu2014towards}, defensive distillation \cite{papernot2016limitations}, protection  against adversarial attacks using generative models \cite{samangouei2018defense}\cite{shen2017ae} focuses on the defense against adversarial attacks. In this work, our aim is to understand the sources and causes of misclassification when the neural network is presented with an adversarial example. 

\myparagraph{Interpretability.}
Explaining the output of a decision maker is necessary to build user trust before deploying them into the real world environment, e.g. in applications like finance, autonomous vehicles, and medical imaging etc. Previous work is broadly grouped into two: 1) \textit{model interpretation}, i.e. understanding of model by observing the structure, parameters and neuronal activities of the networks and 2) \textit{instance level interpretation or prediction explanation}, i.e. showing the causal relationship between input and the specific output \cite{du2018techniques}. De-convolutional neural networks~\cite{zeiler2014visualizing} and activation maximization \cite{simonyan2013deep} fall under the first group. On the other hand, visualizing the evidence for classification \cite{zintgraf2017visualizing}, adding perturbation in the optimization framework and learning the perturbation mask to understand the contribution of features \cite{fong2017interpretable} lie in the second group. As an alternative to visualizations, text-based class discriminative explanations~\cite{hendricks2016generating,park2016attentive} and text-based interpretation with semantic information~\cite{dong2017improving} have also been proposed to explain network decisions. In this work, we use attributes as a means of prediction explanation.

\myparagraph{Interpretability of Adversarial Examples.} After analyzing neuronal activations of the networks for adversarial examples, \cite{dong2017towards} concluded that the networks learn recurrent discriminative parts of objects instead of semantic meaning. In \cite{jiang2018recent}, the authors proposed a datapath visualization module consisting of the layer level, feature level, and the neuronal level visualizations of the network for clean as well as adversarial images. Finally, in \cite{tao2018attacks}, the authors proposed an attribute steered classification model and compared its output with a standard classifier. If outputs were inconsistent then the image was detected as an adversarial image.  They further argued that the interpretation is closely entangled with the detection. Saliency-based model interpretations has been shown to be fragile to interpret adversarial examples~\cite{ghorbani2017interpretation}, i.e. although the output of the neural network for two inputs is different the saliency maps are identical. Similarly, in \cite{zhang2018interpretable}, authors proposed ACID attacks which change the output of saliency maps without changing the output of the classifier. In \cite{adebayo2018sanity}, authors performed sanity checks on saliency-based methods using randomization tests and found that they do not vary with the change in the data generation process and the model. In our work, we propose to ground class-discriminative attributes via bounding boxes that explain class predictions for clean as well as adversarial examples.

\section{Predicting and Grounding Attributes Model}
\begin{figure}[t]
    \centering
        \includegraphics[width=\linewidth, trim=0 0 0 0, clip]{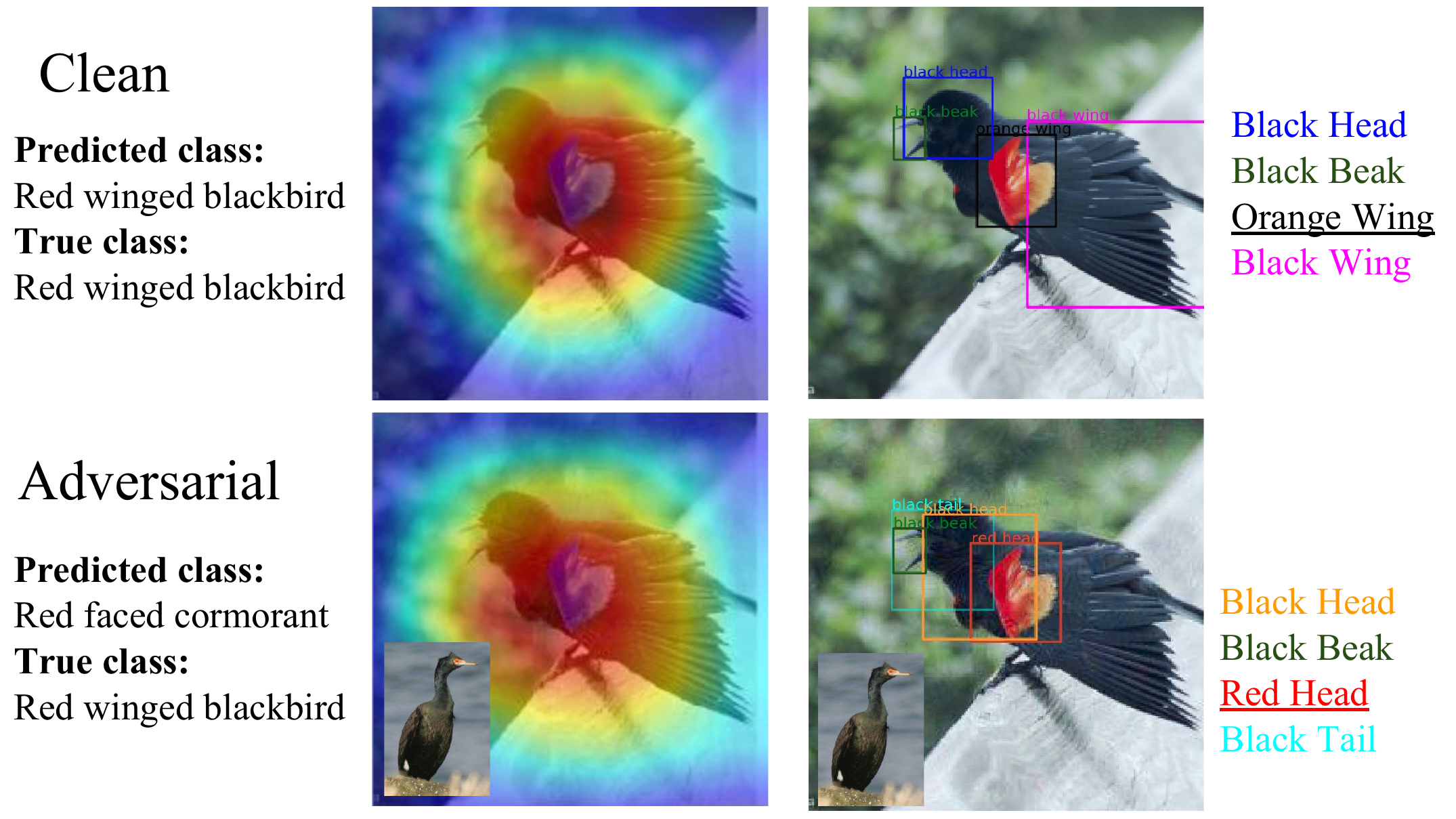}
    \caption{Adversarial images are difficult to explain: when the answer is wrong, often saliency based methods (left) fail to detect what went wrong. Instead, attributes (right) provide intuitive and effective visual and textual explanations.}
   
    \label{fig:saliency}
\end{figure}
Instance level interpretations such as saliency maps~\cite{selvaraju2017grad} are often weak in justifying classification decisions for fine-grained adversarial images, e.g. in Figure~\ref{fig:saliency} the saliency maps of a clean image classified into the correct class, e.g. ``red winged blackbird'', and the saliency map of a misclassified adversarial image, look quite similar. Instead, we propose to predict and ground attributes for both clean and adversarial images to provide visual as well as attribute-based interpretations. In fact, our predicted attributes predicted for clean and adversarial images look quite different. By grounding the predicted attributes one can infer that ``orange wing'' is important for ``red winged blackbird'' while ``red head'' is important for ``red faced cormorant''. Indeed, when the attribute value for orange wing decreases and red head increases the image gets misclassified.

In this section, we detail our two-step framework for interpreting adversarial examples. First, we perturb the images using two different untargeted/targeted adversarial attack methods and robustify the classifiers via adversarial training. Second, we predict class-specific attributes and visually ground them on the image to provide an intuitive justification of why an image is classified as a certain class.

\begin{figure*}
    \centering
    \includegraphics[width=\linewidth]{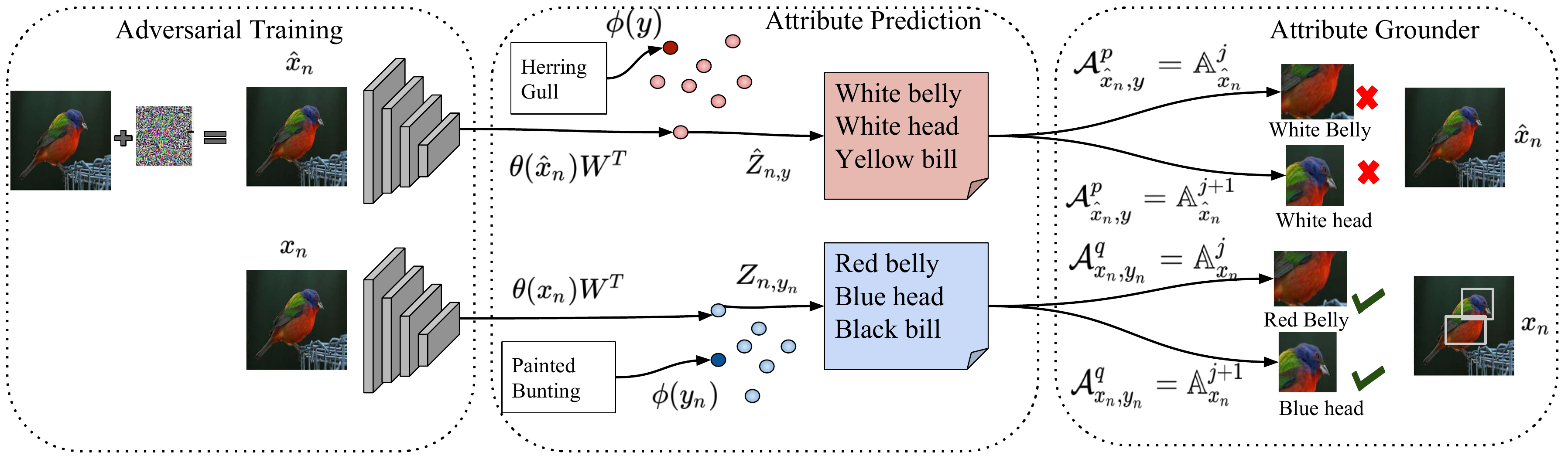}
    \caption{Our interpretable attribute prediction-grounding model. After adversarial attack or adversarial training step, image features of both clean $\theta(x_n)$ and adversarial images $\theta(\hat{x})$ are extracted using Resnet and mapped into attribute space $\phi(y)$ by learning the compatibility function $F(x_n,y_n;W)$ between image features and class attributes. Finally, attributes predicted by SJE $\mathcal{A}_{x_n,y_n}^q$ are grounded by matching them with attributes predicted by Faster RCNN $\mathbb{A}_{x_n}^j$ for clean and adversarial images.}
    \label{fig:ADV_SJE}
   
\end{figure*}    

\subsection{Adversarial Attacks}
We study both untargeted and targeted attacks. Given an original input $x_n$ and its respective correct class $y_n$ predicted by a model $f(x_n)$, an untargeted adversarial attack model generates an image $\hat{x}_n$ for which the predicted class is $f(\hat{x}_n) \neq y_n$. In targeted attacks, for every image $\hat{x}_n$, the the adversary aims at letting the model predict a specific $y_t \neq y_n$. In the following, we detail an adversarial attack method fooling a softmax classifier and an adversarial training technique that robustifies it. 
    
\myparagraph{IFGSM.}
The iterative fast gradient sign method~\cite{kurakin2016adversarial} is a modification of fast gradient sign method (FGSM)~\cite{43405}. In IFGSM, FGSM is applied iteratively solving the objective function to produce adversarial examples:
    \begin{align}
        & \hat{x}_0 =x_n \nonumber \\
        & \hat{x}_n^{i+1}=\text{Clip}_{\epsilon}\{\hat{x}_n^{i}+\alpha\text{Sign}(\bigtriangledown_{\hat{x}_n^i}J(\hat{x}_n^i,y_{n}))\}
    \end{align}
where $\bigtriangledown_{\hat{x}_n^i}J$ represents the gradient of the cost function w.r.t. perturbed image $\hat{x}_n^i$ at step $i$. $\alpha$ determines the step size which is taken in the direction of sign gradient and finally, the result is clipped by epsilon $\text{Clip}_{\epsilon}$.

\myparagraph{Adversarial Training.} As a defense against adversarial attacks~\cite{43405} adversarial training minimizes the objective:
    \begin{align}
        J_{adv}(\theta(x_n),y_n) & = \alpha J(\theta(x_n),y_n) \nonumber \\
        & + (1-\alpha)J(\theta(\hat{x}_n),y)
    \end{align}
where, $\theta(x_n)$ are input image features, $J(\theta(x_n),y_n)$ is the classification loss for clean images, $J(\theta(\hat{x}_n),y)$ is the loss for adversarial images and $\alpha$ regulates the loss to be minimized. The model finds the worst case perturbations and fine tunes the network parameters to reduce the loss on perturbed inputs. Hence, the classification accuracy on adversarial images increases, however there is trade-off between the accuracy of the predictions in clean and adversarial images.
Adversarial training helps in learning more robust classifiers by suppressing the perturbations from adversarial images~\cite{tsipras2018robustness}. 
Further, it is also considered as a regularization technique~\cite{miyato2015distributional}. 
    
\subsection{Attribute Prediction and Grounding}
Our attribute prediction and grounding model uses attributes as side information to define a joint embedding space that the images are mapped to. In this space, attributes act as side information to interpret the classification decision. As shown in Fig.\ref{fig:ADV_SJE}, during training our model maps clean training images close to their respective class attributes, e.g. ``painted bunting'' with attributes ``red belly, blue head, black bill'', whereas adversarial images get mapped close to a wrong class, e.g. ``herring gull'' with attributes ``white belly, white head, yellow bill''. Finally, we visualize the predicted attributes for clean and adversarial images using a pre-trained Faster RCNN model.

\myparagraph{Attribute prediction.} We employ structured joint embeddings (SJE)~\cite{akata2015evaluation} to predict attributes in an image. Given input image features $\theta(x_n) \in \mathcal{X}$ and output class embedding $\phi(y_n) \in \mathcal{Y}$ from the sample set $\mathcal{S}=\{(\theta(x_n),\phi(y_n),n=1...N \}$  SJE learns a mapping $f:\mathcal{X} \to \mathcal{Y}$ by minimizing the empirical risk of the form $\frac{1}{N}\sum_{n=1}^N \Delta(y_n,f(x_n))$ where $\Delta: \mathcal{Y} \times \mathcal{Y} \to \mathbb{R} $ estimates the cost of predicting $f(x_n)$ when the true label is $y_n$.
    
A compatibility function $F:\mathcal{X}\times\mathcal{Y}\to \mathbb{R}$ is defined between input $\mathcal{X}$and output $\mathcal{Y}$ space:
\begin{equation}
        F(x_n,y_n;W)=\theta(x_n)^TW\phi(y_n)
\end{equation}

where $W$ is a matrix of dimension $D\times E$ where $D$ is the dimension of input and $E$ is the dimension of output embedding. It denotes the model parameters to be learned by ranking the correct class higher than the other classes:
    \begin{equation}
        \frac{1}{N}\sum_{n=1}^N\underset{y\in \mathcal{Y}}{\mathrm{max}}\{0,l(x_n,y_n,y)\}
    \end{equation}
where $l(x_n,y_n,y)$ is the pairwise ranking loss:
    \begin{equation}
        \Delta(y_n,y)+\theta(x_n)^TW\phi(y_n)-\theta(x_n)^TW\phi(y)
    \end{equation}
We optimize $W$ with SGD by sampling $(x_n,y)$ and searching for the highest ranked class $y_n$. If the sampled label $y$ is not the correct label then the weights are updated using:
    \begin{equation}
        W^t=W^{t-1}+\eta_t \theta(x_n)[\,\phi(y_n)-\phi(y)]\,^T
    \end{equation}
where $\eta$ is the learning rate and $\theta(x_n)W$ gives the predicted attributes for image $x_n$. The image is assigned to the label of the nearest per-class output embedding $\phi(y_n)$.

\myparagraph{Attribute grounding.} In our final step, we ground the predicted attributes on to the input images using a pre-trained Faster RCNN network and visualize them as in~\cite{anne2018grounding}. The pre-trained Faster RCNN model $\mathcal{F}(x_n)$ predicts bounding boxes denoted by $b^j$. For each object bounding box it predicts the class $\mathbb{Y}^j$ as well as the attribute $\mathbb{A}^j$ ~\cite{anderson2018bottom}.
\begin{equation}
    b^j,\mathbb{A}^j,\mathbb{Y}^j=\mathcal{F}(x_n)
\end{equation}

The most discriminative attributes predicted by SJE are selected based on the criteria that they change the most when the image is perturbed with noise. Then we look up for these attributes $\mathcal{A}_{x_n,y_n}^q, \mathcal{A}_{\hat{x}_n,y}^p$ in attributes predicted by Faster RCNN for each bounding box $\mathbb{A}_{x_n}^j, \mathbb{A}_{\hat{x}_n}^j$ and when the attributes predicted by SJE and Faster RCNN match i.e. $\mathcal{A}_{x_n,y_n}^q = \mathbb{A}_{x_n}^j$, $\mathcal{A}_{\hat{x}_n,y}^p = \mathbb{A}_{\hat{x}_n}^j$ we ground them on their respective clean and adversarial images. Where, $q$ and $p$ are the indexes of the attributes predicted by SJE which change the most when perturbed with adversarial noise.

\section{Experiments}
\begin{figure*}[t]
        \centering
        
            \includegraphics[width=0.162\linewidth, trim=15 0 45 20, clip]{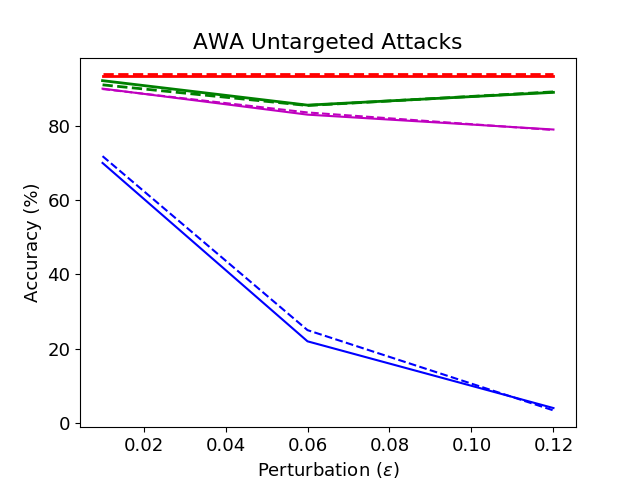}
            \includegraphics[width=0.162\linewidth, trim=15 0 45 20, clip]{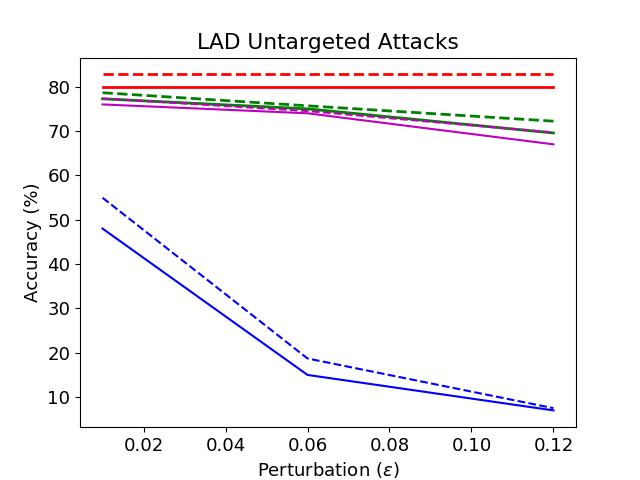}
            \includegraphics[width=0.162\linewidth, trim=15 0 45 20, clip]{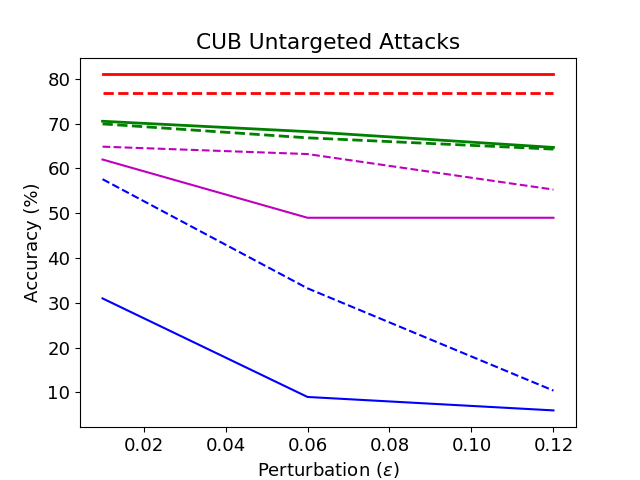} 
            \includegraphics[width=0.162\linewidth, trim=15 0 45 20, clip]{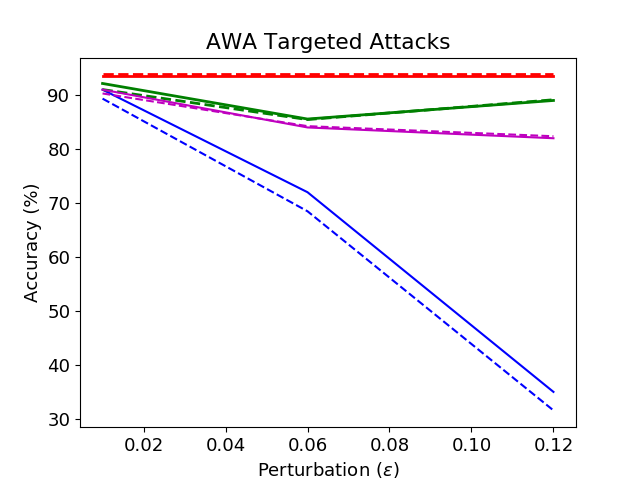}
            \includegraphics[width=0.162\linewidth, trim=15 0 45 20, clip]{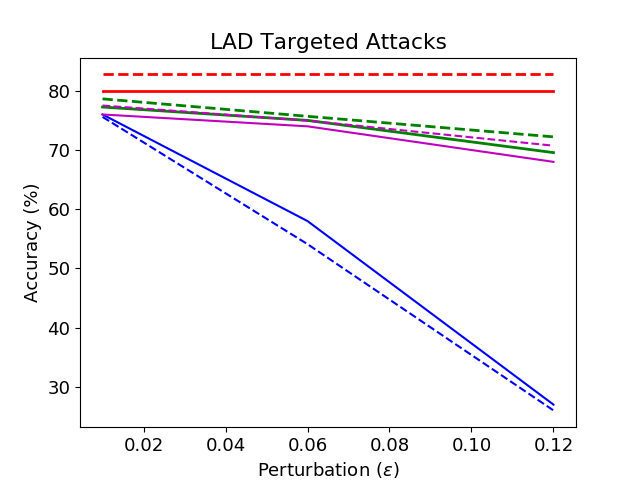}
            \includegraphics[width=0.162\linewidth, trim=15 0 45 20, clip]{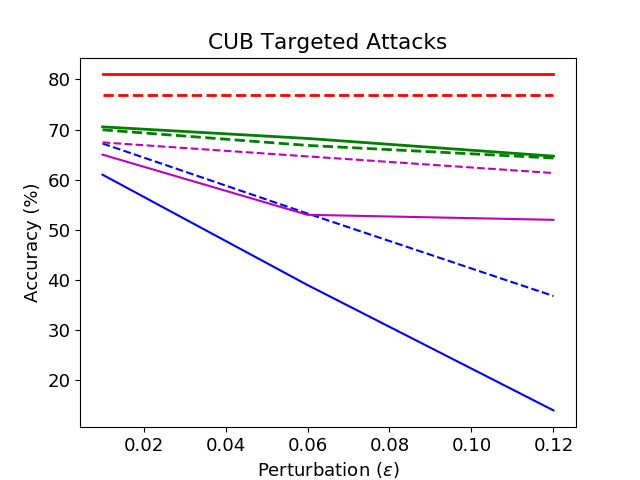}
            \includegraphics[width=0.96\linewidth, trim =0 0 0 0, clip]{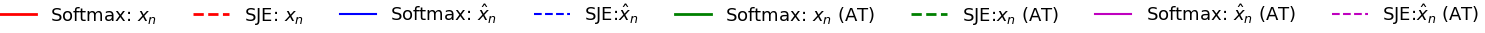}
            
        \caption{Comparing the accuracy of the non explainable Softmax classifier and the explainable SJE classifier for clean and adversarially perturbed samples. We evaluate both classifiers on clean $x_n$ and adversarial images with no adversarial training $\hat{x}_n$ and the same with adversarial training $x_n$ (AT) and $\hat{x}_n$ (AT) respectively.}
        \label{fig:targetedUntargeted}
    \end{figure*}
    
In this section, we perform experiments on three different datasets and analyze model performance for clean as well as adversarial images. Finally, we present quantitative as well as qualitative analysis using attributes for both targeted and untargeted attacks. 

\myparagraph{Datasets.} We experiment on three datasets, i.e. Animals with Attributes 2 (AwA) \cite{lampert2009learning}, Large attribute (LAD) \cite{zhao2018large} and Caltech UCSD Birds (CUB)  \cite{wah2011caltech}. AwA contains 37322 images (22206 train / 5599 val / 9517 test) with 50 classes and 85 attributes per class. LAD has 78017 images (40957 train / 13653 val / 23407 test) with 230 classes and 359 attributes per class. CUB consists of 11,788 images (5395 train / 599 val / 5794 test) belonging to 200 fine-grained categories of birds with 312 attributes per class. 
    
\myparagraph{Image Features and Adversarial Examples.} We extract image features and generate adversarial images using fine-tuned Resnet-152. Our untargeted and targeted attacks using iterative fast gradient sign method with epsilon $\epsilon$ values $0.01$, $0.06$ and $0.12$ and $\l_\infty $ norm as a similarity measure between clean input and the generated adversarial example. We performed targeted attacks under average case scenario where we selected the target class randomly from labels \cite{carlini2017towards}. 
 
As for adversarial training, we repeatedly computed the adversarial examples while training and fine-tuned the Resnet-152 to minimize the loss on these examples. We generated adversarial examples using projected gradient descent method which is a multi-step variant of FGSM with epsilon $\epsilon$ values $0.01$, $0.06$ and $0.12$ respectively for adversarial training as in~\cite{madry2017towards}.

\myparagraph{Attribute Prediction and Grounding.}
Our per-class attribute vectors come with the dataset and are annotated manually. At test time the image features are projected onto the attribute space and the image is assigned with the label of the nearest ground truth attribute vector. 

The predicted attributes are grounded by using Faster-RCNN pre-trained on Visual Genome Dataset since we do not have ground truth part bounding boxes for any of our datasets. The Faster-RCNN model extracts the bounding boxes using 1600 object and 400 attribute annotations. Each bounding box is associated with an attribute followed by the object, e.g. a brown bird.

\subsection{Comparing Softmax and SJE for Classification}
Here, we evaluate Softmax and SJE  classifiers in terms of the classification accuracy on both clean and adversarial images generated with untargeted and targeted attacks for all three datasets. Since SJE model is a more explainable classifier, e.g. predicts attributes, compared to softmax, e.g. predicts directly the class label, it is important to see if there is any significant drop in accuracy. Note that we are not attacking the SJE network directly but we are applying black box attacks on SJE. Similarly, the adversarial training is also performed on Softmax classifier and then the features extracted from this model are used for training SJE.

We observe from our results with targeted and untargeted IFGSM attacks in Figure~\ref{fig:targetedUntargeted} that SJE and Softmax accuracies are on par for clean images on AWA dataset, SJE accuracy is slightly higher for LAD dataset and slightly lower for CUB dataset (red curves). With untargeted adversarial attacks, SJE works slightly better for AWA and LAD datasets and significantly better for CUB dataset i.e. $\approx 30\%$ for $\epsilon=0.025$ (blue curves). However, with targeted attacks for AWA and LAD datasets, SJE accuracy is slightly lower than Softmax but the difference is not significant and is significantly better for CUB dataset (blue curves). This shows that while softmax classifier works slightly better on clean images, SJE works significantly better especially when the perturbation is small and the dataset is fine-grained with well-defined attributes. This shows that when the image is perturbed, by predicting attributes the model not only provides an explanation to the user but also the class predictions are more accurate.  

In addition, for targeted vs untargeted attacks, the accuracy for targeted attacks does not decrease as much as with untargeted attacks on all the three datasets. The reason behind lack in the drop of accuracy for targeted attacks is that in targeted attacks we randomly target the images into wrong class which could be very far from its ground truth hence it becomes difficult to misclassify into targeted class as compared to untargeted attacks where the image gets misclassified into the nearest wrong class.  Although the drop in accuracy for untargeted attacks is higher than targeted attacks (blue curves), but then the improvement in accuracy for untargeted is also high which leads to almost same adversarially robust accuracy for both targeted and untargeted attacks (purple curves).

Our evaluation with and without adversarial training shows that the classification accuracy improves for adversarial images when adversarial training is used. For example for AWA the accuracy improved from $\approx 5\%$ to $\approx 83\%$ for untargeted attack with $\epsilon=0.12$. However, the accuracy for clean images dropped e.g. for AWA the accuracy dropped from $\approx 93\%$ to $\approx 89\%$ for untargeted attack with $\epsilon=0.12$ (green curves). Overall we observe with both targeted and untargeted attacks that SJE is more robust to the adversarial attacks as compared to Softmax (dotted blue curves). Moreover, SJE results improve significantly for adversarial examples as compared to Softmax with adversarial training (dotted purple curves).

\subsection{Quantifying Effect of Predicted Attributes}

 \begin{figure*}[t]
        \centering
        \includegraphics[width=\linewidth, trim=0 0 0 0, clip]{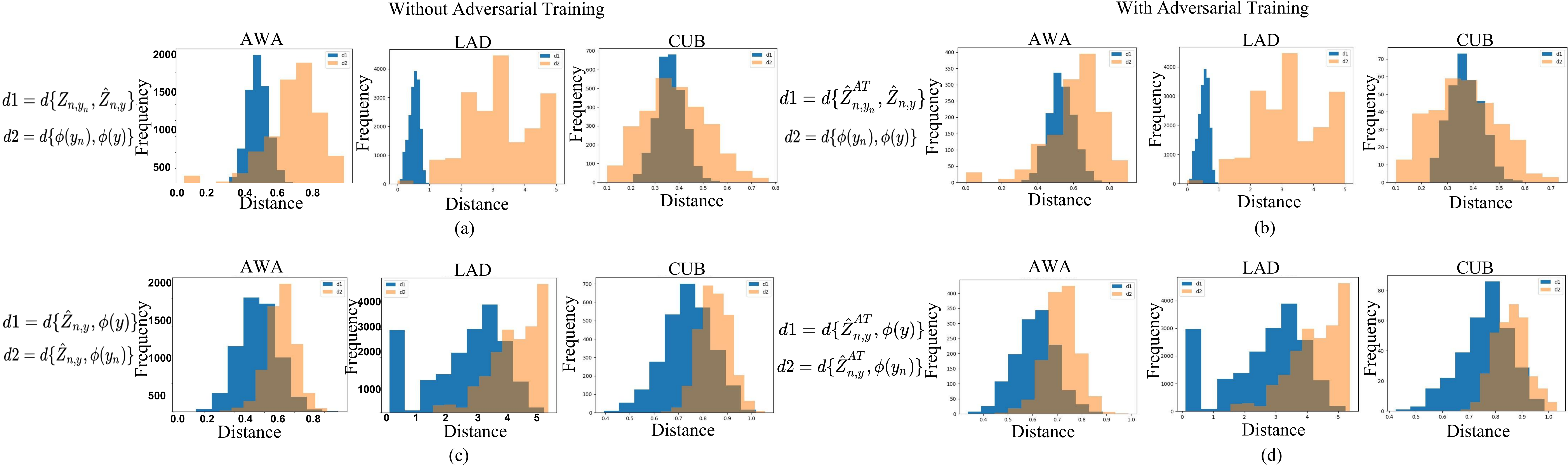}
        \vspace{-3mm}
        \caption{ Attribute distance plots for standard and robust learning frameworks. Standard learning framework plots are shown for clean and adversarial image attributes and robust learning framework plots are shown only for adversarial image attributes but for adversarial images misclassified with standard features and correctly classified with robust features.}
            \label{fig:attributeanalysis}
            \vspace{-3mm}
\end{figure*}

Our aim is to analyze (1) the predicted attributes of the clean images classified correctly $Z_{n,y_n}$, and adversarial images misclassified $\hat{Z}_{n,y}$ without adversarial training (2) predicted attributes of the adversarial images classified correctly $\hat{Z}^{AT}_{n,y_n}$ and classified incorrectly $\hat{Z}_{n,y}^{AT}$ with adversarial training. Note that, the correct ground truth class attribute is referred to as $\phi(y_n)$ and incorrect class attributes are as $\phi(y)$.  
   
We select top $20\%$ of the attributes whose value changes the most with adversarial perturbations considering distances between predicted attributes of clean and adversarial images when they are correctly and incorrectly classified.

We contrast the Euclidean distance between predicted attributes of (correctly classified) clean and (incorrectly classified) adversarial samples:
\begin{equation}
    d_1 = d\{Z_{n,y_n},\hat{Z}_{n,y}\} =\parallel Z_{n,y_n}-\hat{Z}_{n,y} \parallel_2
    \label{eq:d1_1}
\end{equation}
with the Euclidean distance between the ground truth attribute vector of the correct and incorrect classes:
\begin{equation}
   d_2 = d\{\phi(y_n),\phi(y)\}=\parallel\phi(y_n)-\phi(y)) \parallel_2
   \label{eq:d2_1}
\end{equation}
and show the results in Figure~\ref{fig:attributeanalysis} (a). We observe that for AWA and LAD datasets the distances between the predicted attributes for adversarial and clean images $d_1$ are smaller than the distances between the ground truth attributes of clean and adversarial classes $d_2$. This result shows that, only a minimal change in attribute values towards the wrong class can cause a misclassification. On the other hand, the fine-grained CUB dataset behaves differently. The overlap between $d_1$ and $d_2$ distributions shows that the images from fine-grained classes are more susceptible to adversarial attacks and hence their attributes change significantly compared to images of coarse categories. 
        
Contrasting the distances between the predicted attributes for the adversarial image $\hat{Z}_{n,y}$ and the ground truth attribute $\phi(y)$ for the adversarial class:  
        \begin{equation}
            d_1 = d\{\hat{Z}_{n,y},\phi(y)\}=\parallel \hat{Z}_{n,y}-\phi(y) \parallel_2
            \label{eq:d1_2}
        \end{equation}
with the distance between the predicted attribute for adversarial image $\hat{Z}_{n,y}$ and the ground truth attribute $\phi(y_n)$ for the correct class:  
\begin{equation}
    d_2 = d\{\hat{Z}_{n,y},\phi(y_n)\}=\parallel \hat{Z}_{n,y}-\phi(y_n) \parallel_2
    \label{eq:d2_2}
\end{equation}
we obtain results in Figure~\ref{fig:attributeanalysis} (c). We observe that, for most of the images, the distance between the adversarial image attribute and the correct class attribute $d2$ is higher than the distance between adversarial image attribute and the wrongly classified class $d1$. This result shows us that, adversarial images are misclassified because the adversarial image attributes are close to the incorrect class attributes whereas they are far away from the correct class attributes.

\begin{figure*}[t]
    \centering
    \includegraphics[width=\linewidth, trim=0 0 0 0, clip]{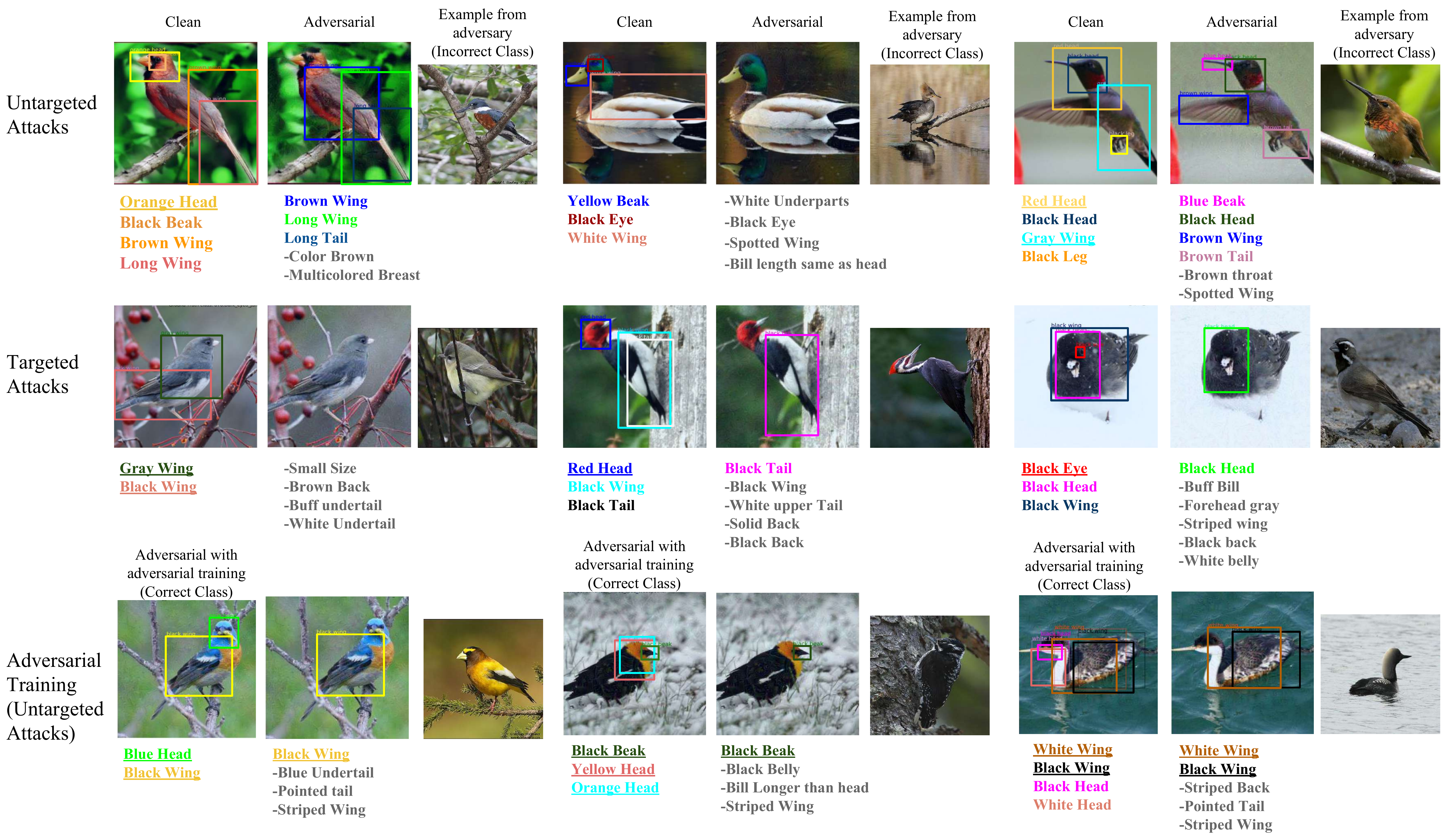}
    \vspace{-5mm}
    \caption{Qualitative analysis for  untargeted/targeted attacks and adversarial training on CUB. The attributes ranked by importance for the classification decision are shown below the images. The grounded attributes are color coded for visibility (the ones in gray could not be grounded). The attributes for clean images (and adversarial images with adversarial training) are related to correct classes whereas the ones predicted for adversarial images change towards incorrect classes. }
    \label{fig:Qualitative-1}
    \vspace{-3mm}
\end{figure*} 

Our results comparing the distances between the predicted attributes of the adversarial images that are classified correctly with the help of adversarial training $\hat{Z}^{AT}_{n,y_n}$ and incorrectly without adversarial training $\hat{Z}_{n,y}$:
\begin{equation}\label{eq:d1_3}
    d_1 = d\{\hat{Z}^{AT}_{n,y_n},\hat{Z}_{n,y}\}=\parallel \hat{Z}^{AT}_{n,y_n}-\hat{Z}_{n,y} \parallel_2
\end{equation}
with the distances between the ground truth target class attributes $\phi(y_n)$ and ground truth wrong class attributes $\phi(y)$:
\begin{equation}\label{eq:d2_3}
    d_2 = d\{\phi(y_n),\phi(y)\}=\parallel\phi(y_n)-\phi(y)) \parallel_2
\end{equation}
are shown in Figure~\ref{fig:attributeanalysis} (b). We observe that the overall behavior of the predicted attributes for that adversarial images with adversarial training and without adversarial training is similar to the behavior seen in Figure~\ref{fig:attributeanalysis} (a) for clean and adversarial images. This shows that the adversarial images with adversarial training behave like clean images, i.e. predicted attributes for the adversarial images with adversarial training become closer to their ground truth correct class. 
        
We compare the distances between the predicted attributes for the incorrectly classified adversarial image $\hat{Z}^{AT}_{n,y}$ and the ground truth attribute $\phi(y)$ of the adversarial class: 
\begin{equation}
    d_1 = d\{\hat{Z}^{AT}_{n,y},\phi(y)\}=\parallel \hat{Z}^{AT}_{n,y}-\phi(y) \parallel_2
    \label{eq:d1_2}
\end{equation}
with the distance between the predicted attribute for adversarial image $\hat{Z}^{AT}_{n,y}$ and the ground truth attribute $\phi(y_n)$ for the correct class:
\begin{equation}
    d_2 = d\{\hat{Z}^{AT}_{n,y},\phi(y_n)\}=\parallel \hat{Z}^{AT}_{n,y}-\phi(y_n) \parallel_2
    \label{eq:d2_2}
\end{equation}
when the classifier is trained with adversarial training. From the results in Figure~\ref{fig:attributeanalysis} (d) we observe a similar behavior as the results presented in Figure~\ref{fig:attributeanalysis} (c). This shows adversarial images misclassified with adversarial training behaves like adversarial images misclassified without it.

\subsection{Grounding Predicted Attributes}

To qualitatively analyse the predicted attributes, we ground them on clean and adversarial images. We select our images among the ones that are correctly classified when clean and incorrectly classified when adversarially perturbed. For clean images (or adversarial images with adversarial training), we select the most discriminative attributes based on:
\begin{equation}
        q=\underset{i}{\mathrm{argmax}}(Z_{n,y_n}^i-\phi(y^i))
        \label{eq:att_sel1}
\end{equation}
for adversarial images we select them based on: 
\begin{equation}
        p=\underset{i}{\mathrm{argmax}}(\hat{Z}_{n,y}^i-\phi(y_n^i)). 
        \label{eq:att_sel2}
\end{equation}
We evaluate $50$ attributes that change their value the most for CUB, $50$ attributes for AWA, and  $100$ attributes for LAD dataset. We match the selected attributes with the attributes predicted by Faster RCNN to ground them on the images.

\begin{figure*}[t]
    \centering
    \includegraphics[width=\linewidth, trim=0 0 0 0, clip]{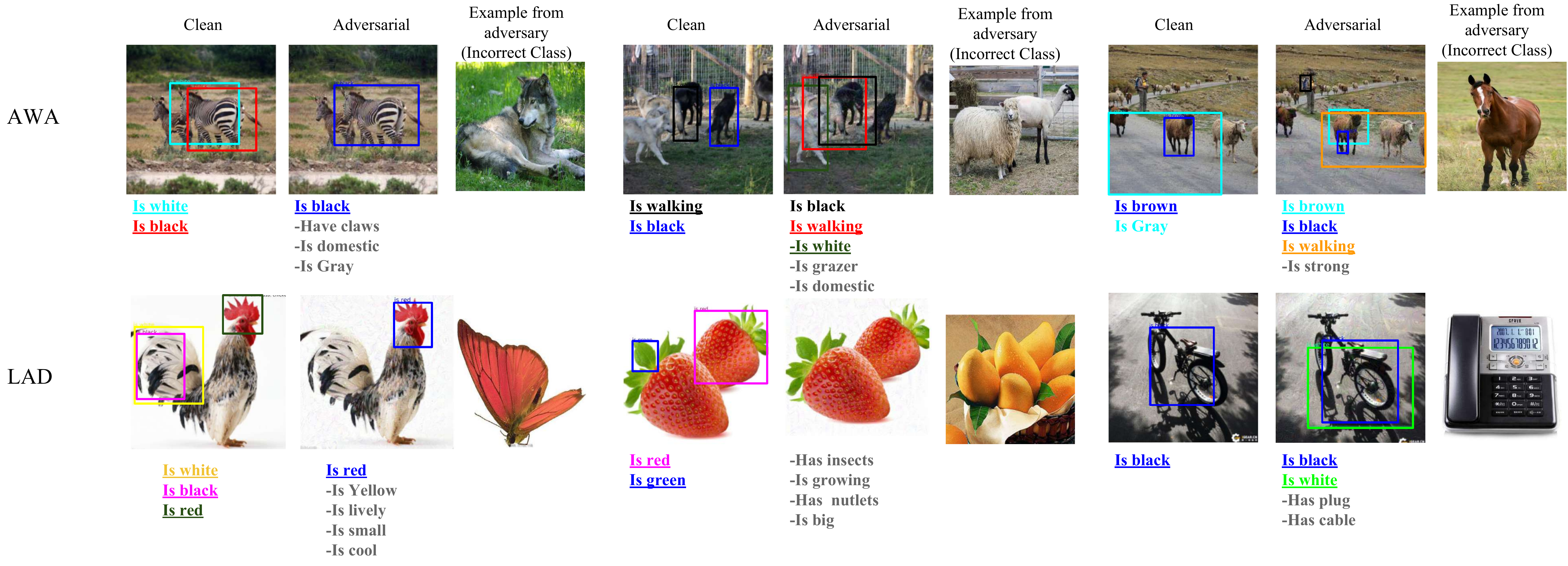}
    \vspace{-8mm}
    \caption{Qualitative analysis for untargeted attacks on AWA and LAD. The attributes are ranked by importance for the classification decision, the grounded attributes are color coded for visibility (the ones in gray could not be grounded). }
    \label{fig:Qualitative-2}
    \vspace{-3mm}
\end{figure*}
   
\myparagraph{Qualitative Results in CUB.}
We perform an analysis with untargeted and targeted IFGSM attacks as well as adversarial training on the fine-grained CUB dataset. 

In untargeted attacks (results at the top row of Figure~\ref{fig:Qualitative-1}), the image gets misclassified into the nearest incorrect class. We observe that the most important attributes for the clean images are localized accurately; however, for adversarial images misclassifications occur. Those attributes which are common among both clean and adversarial classes are localized correctly on the adversarial images; however, the attributes which are not related to the correct class, i.e. the ones that are related to the wrong class can not get grounded as there is no visual evidence that supports the presence of these attributes. For example ``brown wing, long wing, long tail'' attributes are common in both classes; hence, they are present both in the clean image and the adversarial image. On the other hand, has a brown color and a multicolored breast which are evidences that are not present in the adversarial image. Hence, they can not get grounded. Similarly, in the second example none of the attributes are grounded. This is because the evidence for those attributes are not present in the image. In the third example, common attributes are localized but ``brown throat, spotted wing'' are not localized for the same reasons.

In targeted attacks (results in the middle row of Figure~\ref{fig:Qualitative-1}) the images are forced to get missclassified into a randomly selected class. So, the images get missclassified into classes that do not share many common attributes. Our first visualizations show that none of the attributes of the adversary class were visible in the adversarial example, hence, those attributes could not get grounded. In other words, predicted adversarial image attributes are in accordance with the wrong class attributes but different from the clean image so none of the attribute got localized. For the second image, black tail is a common property between the clean image class and the adversary however, this is not the most discriminating property. One of the most discriminating properties such as ``solid back'' did not get localized since there is no visual evidence that supports the presence of this attribute in clean image. Similarly, in the third example, we observe that the most discriminating property is ``striped wing'' but it did not get localized in the adversarial image for the same reason.
    
Finally, our analysis with correctly classified images due to adversarial training shows that adversarial images with adversarial training behave like clean images also visually. In last row of Figure~\ref{fig:Qualitative-1}, we observe that the attributes of adversarial image without adversarial training are closer to the adversarial class attributes. However, the grounded attributes of adversarial image with adversarial training are closer to its ground truth class. For instance, the first example contains a ``blue head'' and a ``black wing'' whereas one of the most discriminating properties of the correct class ``blue head'' is not relevant to the adversarial class hence this attribute is not predicted as the most relevant by our model and hence our attribute grounder did not ground it.

\myparagraph{Qualitative Results in AWA and LAD.} Due to restricted space, we provide results on AWA and LAD only with images perturbed with untargeted attacks. Our results in Figure~\ref{fig:Qualitative-2} show that the grounded attributes on clean images conform the classification into the correct class while the attributes grounded on adversarial images are common among clean and adversarial images. For instance first example of AWA ``is black'' attribute is common in both classes so it is grounded on both images but ``has claws'' is an important attribute for the adversarial class. As it is not present in correct class, it is not grounded. 

On the other hand, compared to misclassifications caused by adversarial perturbations on CUB, as AWA and LAD are coarse grained datasets, images do not necessarily get misclassified into the most similar class. Therefore, there is less overlap of attributes between correct and adversarial classes, which is in accordance with our quantitative results. Furthermore, the attributes for both datasets are not highly structured as different objects can be distinguished from each other with only a small number of attributes. Our method grounds the common attributes. The second example for LAD in Figure~\ref{fig:Qualitative-2} shows that attributes such as ``red'' and ``green'' are distinguishing for ``strawberry'' which are correctly predicted and grounded. On the other hand, ``has nutlets'' and ``is big'' are attributes distinguishing for ``Mango''; hence, they can not be grounded on the adversarially perturbed strawberry image. 

\section{Conclusion}
In this work, we proposed an attribute prediction and grounding framework to explain why adversarial perturbations cause misclassifications. Our model predicts class-specific properties of the objects via ranking relevant class attributes higher than irrelevant ones and grounds these attributes on their respective images. Our analysis involved images generated by targeted and untargeted attacks as well as adversarial training. We showed quantitatively and qualitatively that predicted attributes for adversarial images are relevant to the wrong class and to the correct class for clean images justifying why adversarial images get misclassified. We visually grounded these predicted attributes to show the visible and missing evidence when a misclassification occurs on three benchmark datasets.

\end{document}